\documentclass[letterpaper, 10 pt, conference]{ieeeconf}

\IEEEoverridecommandlockouts                              % This command is only needed if 
\usepackage[caption=false, font=footnotesize]{subfig}
\usepackage{graphicx} % for pdf, bitmapped graphics files
\usepackage{algorithm} 
\usepackage{algpseudocode} 
\usepackage{hyperref}
\usepackage{amsmath} % assumes amsmath package installed
\usepackage{amssymb}
\usepackage[utf8]{inputenc}
\usepackage{array}
\usepackage{booktabs}
\usepackage{tabularx}
\usepackage{ragged2e}
\usepackage{adjustbox}
\usepackage{siunitx}
\usepackage{subfig}
\usepackage{cite}
\def\BibTeX{{\rm B\kern-.05em{\sc i\kern-.025em b}\kern-.08em
    T\kern-.1667em\lower.2ex\hbox{E}\kern-.125emX}}

\title{\LARGE \bf
Bumper Drone: Elastic Morphology Design for Aerial Physical Interaction
}

\author{Pongporn Supa$^{1}$, Alex Dunnett$^{2}$, Feng Xiao$^{3}$, Rui Wu$^{1}$, Mirko Kovac$^{4}$, and Basaran Bahadir Kocer$^{2}$ % <-this % stops a space
\thanks{*This work was supported by funds from University of Bristol.}% <-this % stops a space
\thanks{$^{1}$School of Engineering Mathematics and Technology, University of Bristol.}
\thanks{$^{2}$School of Civil, Aerospace and Design Engineering, University of Bristol.}
\thanks{$^{3}$Department of Aeronautics, Imperial College London, London, UK.}
\thanks{$^{4}$Laboratory of Sustainability Robotics, EMPA, Dubendorf, Switzerland.}
}

\begin{document}
\bstctlcite{IEEEexample:BSTcontrol}
\maketitle
\thispagestyle{empty}
\pagestyle{empty}

%%%%%%%%%%%%%%%%%%%%%%%%%%%%%%%%%%%%%%%%%%%%%%%%%%%%%%%%%%%%%%%%%%%%%%%%%%%%%%%%
\begin{abstract}

%Aerial robots have shifted their applications from passive observation to active physical interaction with the environment in risk-prone tasks for human workers. Previous research has successfully demonstrated their capabilities in manipulation tasks with minimal attention to tactile navigation. This task typically involves push-and-slide operations affected by friction forces upon contact with unknown surfaces. While many studies focus on handling friction forces, this work takes a different approach by performing repeated, brief contacts called touch-and-go manoeuvres, enabling aerial robots to use tactile feedback for navigation. The challenge in this behaviour lies in the ability to recover from collisions. Inspired by nature, we designed a whisker-inspired drone bumper that passively recovers from collisions and performs such flying behaviour with minimal control. We propose a real-time preprocessing method to mitigate the inherent nonlinearity, hysteresis, and creep effects of our soft whisker sensor design. By combining biomimetic whisker morphology with flying insect reflex-landing behaviour, we demonstrate self-recovery mechanisms where our whisker design acts as an embodied mass-spring-damper system. Experiments show that our bumper drone can absorb impact energy from collisions while maintaining vehicle stability, achieving minimal attitude disturbances with pitch oscillations of only \ang{4} and roll oscillations of just \ang{0.74}. Additionally, the bumper drone can maintain its stability while applying sustained normal contact force with the vertical wall.

Aerial robots are evolving from avoiding obstacles to exploiting the environmental contact interactions for navigation, exploration and manipulation. A key challenge in such aerial physical interactions lies in handling uncertain contact forces on unknown targets, which typically demand accurate sensing and active control. We present a drone platform with elastic horns that enables touch-and-go manoeuvres -- a self-regulated, consecutive bumping motion that allows the drone to maintain proximity to a wall without relying on active obstacle avoidance. It leverages environmental interaction as a form of embodied control, where low-level stabilisation and near-obstacle navigation emerge from the passive dynamic responses of the drone-obstacle system that resembles a mass-spring-damper system. Experiments show that the elastic horn can absorb impact energy while maintaining vehicle stability, reducing pitch oscillations by 38\% compared to the rigid horn configuration. The lower horn arrangement was found to reduce pitch oscillations by approximately 54\%. In addition to intermittent contact, the platform equipped with elastic horns also demonstrates stable, sustained contact with static objects, relying on a standard attitude PID controller.
% achieving minimal attitude disturbances with pitch oscillations of only 3.14 degrees. In addition to intermittent contact, the platform equipped with elastic horns also demonstrates stable sustained contact with static objects, relying on a standard attitude PID controller.

%Additionally, with integrated flex sensors in the horns, the drone can sustain a normal contact force to a vertical wall under manual control. 

\end{abstract}

%%%%%%%%%%%%%%%%%%%%%%%%%%%%%%%%%%%%%%%%%%%%%%%%%%%%%%%%%%%%%%%%%%%%%%%%%%%%%%%%
\section{INTRODUCTION}

Drones are undergoing a paradigm shift from passive observation to active physical interaction with the external environment. This transformation has unlocked new possibilities for aerial physical interaction, particularly in high-risk operations such as inspection and maintenance of wind turbines, bridges, and high-voltage power lines, where human operations are dangerous and time-consuming \cite{AerialManipulationReviewApplication}. By attaching rigid or articulated end-effectors to aerial platforms, researchers have developed aerial manipulators capable of intentionally interacting with the environment to perform diverse manipulation tasks. Recent research has demonstrated successful contact-based tasks across various scenarios, including object pushing \cite{PushingMoving}, grasping \cite{GraspingCan, GraspingDeltaArm}, perching \cite{TreecreeperDrone,TendonPerching}, emergency switch activation \cite{PressingEmergencySwitch}, contact-based inspection \cite{ContactBasedGelSight, ContactBasedLearning, ContactInspectionFT,EmbodiedAPhI}, aerial calligraphy \cite{FlyingCalligrapher} and canopy exploration \cite{romanello2026treespider}.

However, few studies have explored leveraging these physical interaction capabilities for navigation purposes \cite{ContourFollowing, TactileNavigation}. In challenging environments where traditional sensors such as LiDAR or cameras perform poorly due to limited visibility (smoke, fog, dust, or complete darkness), direct environmental contact offers an alternative sensing modality \cite{HapticFeedback}. This approach, known as tactile navigation (TN) \cite{TactileNavigation}, enables aerial manipulators to estimate their location and orientation relative to contacted surfaces, similar to how nocturnal or vision-impaired animals navigate through touch.

\begin{figure}[t!]
    \centering
       \includegraphics[width=\linewidth]{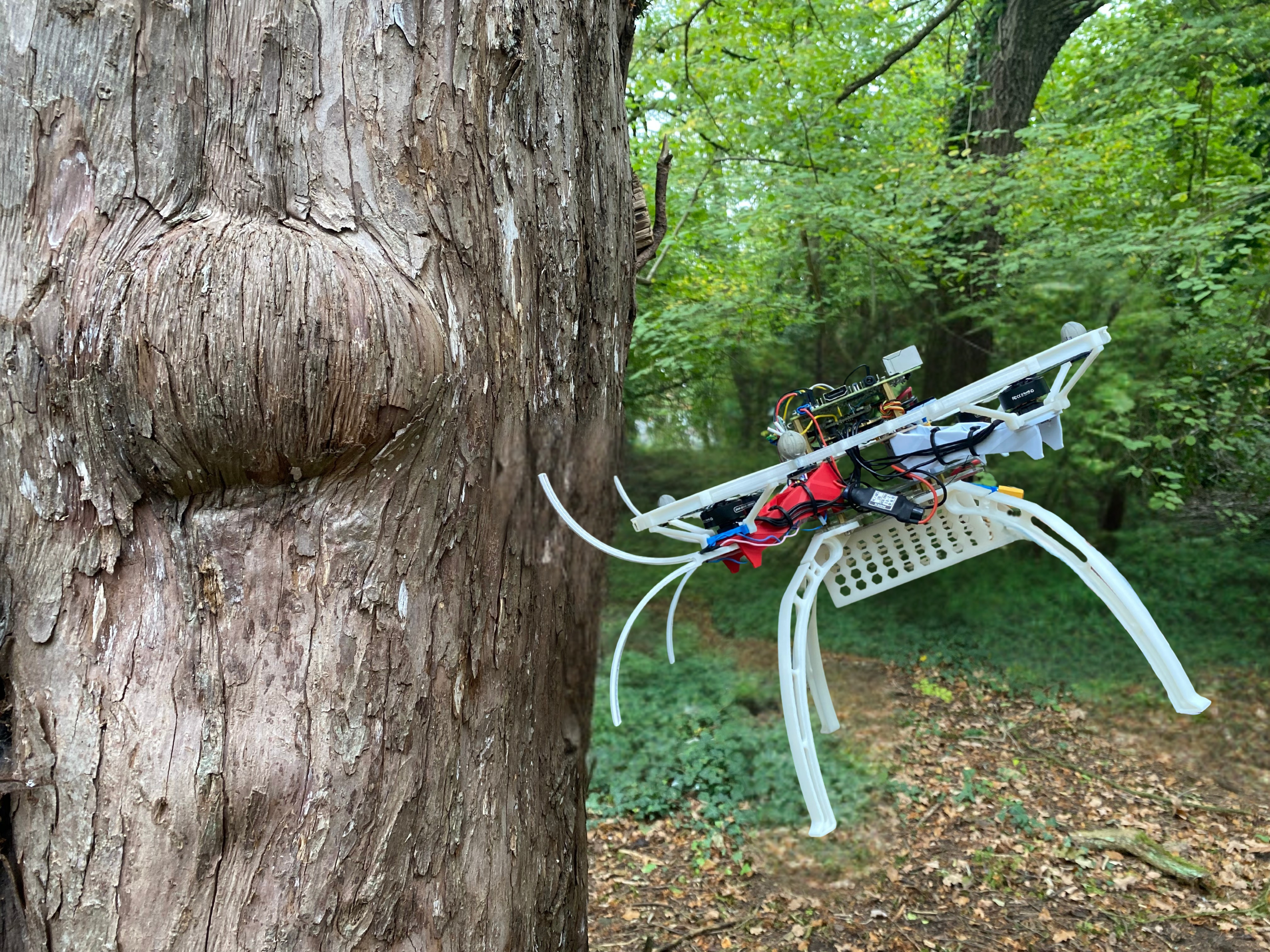}
  \caption{Bumper drone: a \num{700}\unit{\gram} flying robot that has a self-recovery mechanism after collisions and can exert force on objects while maintaining its stability.}
  \label{fig:droneplatform} 
\end{figure}

TN often requires the flying vehicle to push and/or slide the end-effectors on and along the surface. These unknown surfaces vary in friction properties, cannot be directly measured, and introduce uncertainties in contact forces. Current approaches address these challenges through various strategies. In \cite{ContactInspectionFT}, an axis-selective impedance control is proposed to compensate for directional disturbances, including friction forces, by adjusting impedance parameters. A hybrid motion-force controller that incorporates both normal and friction forces into contact wrench models has been studied for time-varying contact force \cite{FlyingCalligrapher}. A learning-based adaptive control strategy for aerial sliding tasks based on an impedance controller using proprioceptive and tactile sensing \cite{ContactBasedLearning}. While effective, these approaches require omnidirectional platforms and precise force-torque (F/T) sensors, at the cost of increased system weight and control complexity~\cite{ContactBasedLearning,ContactInspectionFT,FlyingCalligrapher}. %Alternatively, several studies propose compliant end-effector designs that can assume single-point contact and minimise friction through careful material selection and morphology, such as PLA spherical fingertips \cite{TactileNavigation} and metal ball casters \cite{ContourFollowing}. However, these approaches remain sensitive to small contact force variations that compromise surface orientation estimation accuracy.

% However, few studies have explored leveraging these physical interaction capabilities for navigation purposes \cite{ContourFollowing, TactileNavigation}. In challenging environments where traditional sensors such as LiDAR or cameras perform poorly due to limited visibility (smoke, fog, dust, or complete darkness), direct environmental contact offers an alternative sensing modality. This approach, known as Tactile Navigation (TN) \cite{TactileNavigation}, enables aerial manipulators to estimate their location and orientation relative to contacted surfaces, similar to how nocturnal or vision-impaired animals navigate through touch.

We propose a novel morphology design that minimises the reliance on sensing and active control through a touch-and-go manoeuvre similar to what houseflies do. When they collide with unavoidable obstacles using their heads, they extend their legs by reflex to decelerate and dampen the impact through a mechanism known as \textit{reflex-landing}, which provides controlled rebound and post-collision stability \cite{InsectCollision}. Drawing from this biological analogy, our drone platform employs compliant horns that store impact energy from frontal collisions (Figure~\ref{fig:droneplatform}), behaving as a spring-mass-damper system. The passive dynamic response of this system regulates the bumping behaviour, enabling stable, self-sustained consecutive bumping without sophisticated control. 

\begin{figure*}[htb]
    \centering
    {\includegraphics[width=0.8\linewidth]{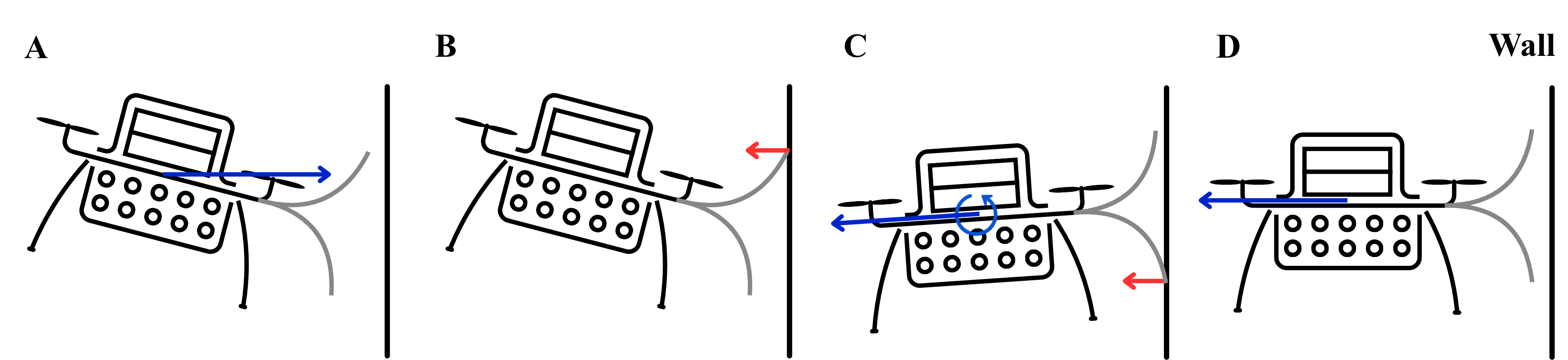}}
    \caption{Drawing of the hypothetical behaviour of our horn design attached to the drone during collision with a vertical wall. (A) The drone approaches the wall with forward velocity and initial pitch angle. (B) Two upper horns make initial contact with the wall. (C) Reaction forces and moments from the wall induce an anticlockwise rotation of the drone, bringing two lower horns into contact. (D) The lower horns produce counter-reaction forces and moments, allowing the drone to rebound and stabilise its attitude passively.}
  \label{fig:hypotheticalbumping} 
\end{figure*}

Furthermore, by integrating resistive thin-film flex sensors into the elastic horns, the platform can estimate normal contact force and displacement based on horn deformation, mimicking biological whisker mechanoreception. Unlike existing whisker-based designs that employ MEMS barometers \cite{WhiskerAerial}, FBG sensors \cite{WhiskerUnderWater}, or Hall effect sensors \cite{lin2022whiskerinspiredtactilesensingcontact} for contact localisation, this design provides measurable contact forces sufficient for object pushing and enhances resilience to collisions.

The present study demonstrates that embodied control can simplify the active control of aerial robots, particularly benefiting miniature platforms with low payload capacity. This approach allows a pushing manoeuvre in which the drone applies a constant normal force to a vertical wall while remaining stable under manual control and without any dedicated control strategies.

\section{HORN DESIGN AND FABRICATION}\label{Whiskerdesign}
\subsection{Conceptual Design}\label{sec:concept}

Our conceptual design integrates two bio-inspired principles: (1) the tactile sensing mechanism of whiskers for contact force and displacement measurement through deformation; (2) the post-collision stabilisation behaviour of houseflies for passive recovery. In biological systems, whiskers provide safe physical interaction with the environment through mechanoreception at their follicle base by transducing forces through follicle deformation \cite{WhiskerAerial}. Inspired by this, we developed curved, flexible horns that naturally transduce contact forces into measurable deformations.

To replicate housefly collision dynamics, four horn arrays (two upper, two lower) are symmetrically arranged in front of the drone platform, providing a distributed contact system that enables passive bumping behaviour without active control. Our collision dynamics hypothesis centres on sequential contact mechanics (Figure~\ref{fig:hypotheticalbumping}). Initial surface contact occurs through the upper horn pair, generating reaction forces and positive pitch moments while also acting as a damping system. Subsequent lower horn engagement acts as another damping system, dissipating energy and providing counteracting moments for passive stabilisation and rebound off the contacted surface. All recovery dynamics are achieved mechanically, without sensor feedback to the attitude controller.

% This design aligns with principles of insect post-collision flight, where body posture stabilisation takes precedence over trajectory control \cite{InsectCollision}. Our experimental validation confirms that collision recovery occurs passively through the inherent morphology of the whisker sensor configuration and pitch angle correction, successfully replicating the collision strategies observed in flying insects.

\begin{figure}[b!]
    \centering
       \includegraphics[width=\linewidth]{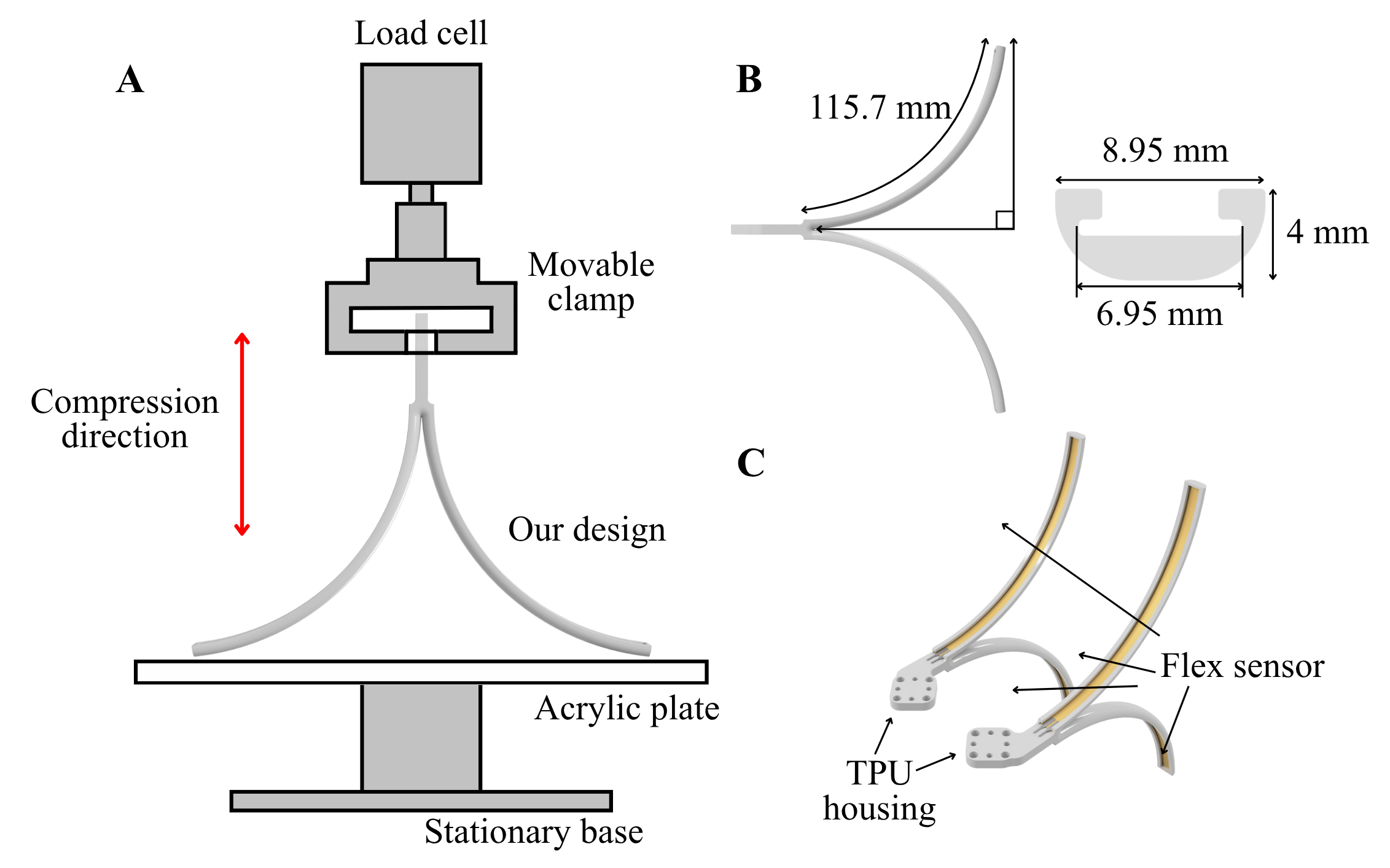}
    \caption{Experimental setup and detailed design. (A) Instron testing machine setup for calibration with compression direction indicated. (B) Side view and cross-section with dimensions. (C) Overview design comprising four flex sensors placed inside left and right 3D-printed TPU housing.}
  \label{fig:horndesign} 
\end{figure}

Our horn design consists of two main parts: 1) Flexible thin-film resistive sensors directly measuring horn deformation for contact force and displacement estimation; 2) a 3D-printed housing providing collision impact energy absorption and passive bumping behaviour (Figure~\ref{fig:horndesign}B, ~\ref{fig:horndesign}C). Our horn housing design is fabricated using a Bambu Lab A1 printer with TPU for AMS (Shore 68D, 0.6mm nozzle, zero infill). The flex sensors are placed inside the 3D-printed housing that is curved to ensure consistent bending direction when making contact with surfaces. Each horn weighs only \num{7}\unit{\gram}.

%In biological systems, whiskers enable safe physical interaction with the environment through mechanoreception at their follicle base. Upon contact, forces and moments transmitted through the whisker shaft cause follicle deformation that mechanoreceptors convert into neural signals for processing and behavioural response. This bio-inspired tactile sensing provides an alternative approach to force measurement during physical interaction that is lightweight, highly sensitive, and eliminates reliance on expensive force-torque sensors whose structural housing requirements significantly increase system weight.

%Many researchers have implemented whisker-based sensing using MEMS barometers \cite{WhiskerAerial}, fibre Bragg grating (FBG) sensors \cite{WhiskerUnderWater}, Hall effect sensors \cite{lin2022whiskerinspiredtactilesensingcontact}, etc., for contact detection. However, these approaches involve delicate manufacturing processes and are primarily designed for contact point localisation where contact forces are insufficient for manipulation tasks. Our work focuses on contact detection and force measurement at the contact point suitable for aerial physical interaction applications, providing a cost-effective sensing solution that is easy to reproduce.

%In traditional whisker sensing, researchers use Nitinol wires as shafts in their designs to minimise friction forces upon interaction \cite{WhiskerAerial, lin2022whiskerinspiredtactilesensingcontact, WhiskerUnderWater}. However, our approach diverges from the conventional sliding-contact approach by implementing a touch-and-go navigation strategy that necessitates robust collision recovery mechanisms inspired by flying insect behaviour. 

\subsection{Horn Characterisation}\label{sec:whiskercharacteristic}

We analysed our horn design morphology to establish the relationship between the resistance of the flex sensors and contact force/displacement using systematic calibration methods. The experimental setup employed an Instron tensile testing machine in compression mode to simultaneously measure force and displacement while pressing horn designs against an acrylic plate (Figure~\ref{fig:horndesign}A).

% Prior testing was performed to find the maximum operational force of \num{10}\unit{\newton} for single-sided horn design (comprising 2 flex sensors) before the horn exhibits irreversible deformation or structural failure.% Since each flex sensor exhibits different resistance readings in its unbent, neutral state (zero deformation), a calibration signal is essential to account for these inherent baseline variations. We collected \num{1000} resistance readings from each sensor at the neutral state, calculated their mean values, and subtracted baseline values from all subsequent resistance readings before each trial. This calibrated baseline resistance is denoted as $\Omega_\text{i, 0}$. 

%Subsequently, we performed whisker levelling by placing thin paper above the acrylic plate and adjusting the height of the whisker sensor placement until the measured force from the load cell exceeded zero. We set this point as our initial reference where both displacement and force were set to zero. 
%Each measurement consisted of loading and unloading phases: compression from \qtyrange{0}{5}{\newton} followed by release back to \num{0} \unit{\milli\metre} displacement, both at identical compression rates. We used zero displacement as the stopping criterion for our experimental trials. Compression rates of \num{1000} \unit{\mm\per\min} was used, with five repetition cycles per rate. 

Each trial consisted of loading from \qtyrange{0}{5}{\newton} and unloading back to \num{0} \unit{\newton} at a compression rate of \num{0} \unit{\milli\metre}/\unit{min}. Five cycles were performed. The reported resistance is calibrated by subtracting the resistance value at the unbent, neutral state. The calibration assumes symmetric behaviour between upper and lower horns, distributing the \num{5} \unit{\newton} maximum force equally (\num{2.5} \unit{\newton} per horn).

\begin{figure}[t!]
    \centering
    \includegraphics[width=\linewidth]{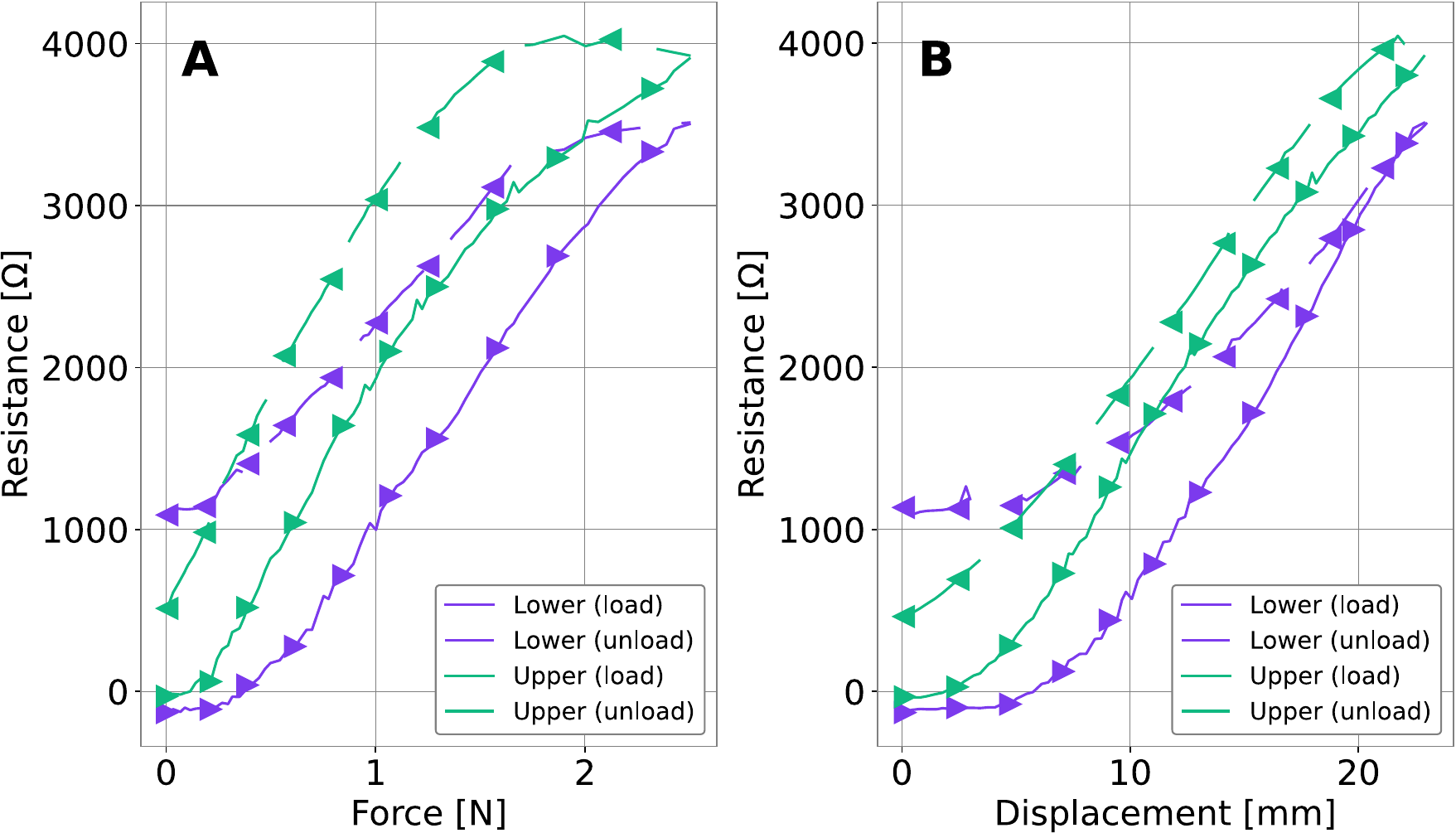}
    \hfill
    % \subfloat[Comparison of preprocessing steps for whisker sensor reading to address effects of vibration, hysteresis, and creep effect. The experimental setup involves a single whisker sensor mounted on a grounded drone operating at 50\% throttle without propellers. The trial consists of two contact events where the drone is manually pushed against a vertical surface and subsequently released. Raw sensor measurements (light gray trace) exhibit baseline noise during the initial period, followed by drone arming and throttle engagement via remote control, two wall contact cycles, and final release. Real-time preprocessing results are displayed following baseline calibration. The purple trace represents a filtering method adapted from literature \cite{WhiskerAerial} (implemented on different hardware). The black trace shows our methodology that integrates hysteresis compensation and dynamic baseline adjustment depending on the features of TPU material during the unloading phase, using rate-of-change thresholds.\label{fig:signalprocessing}]{%
    \includegraphics[width=\linewidth]{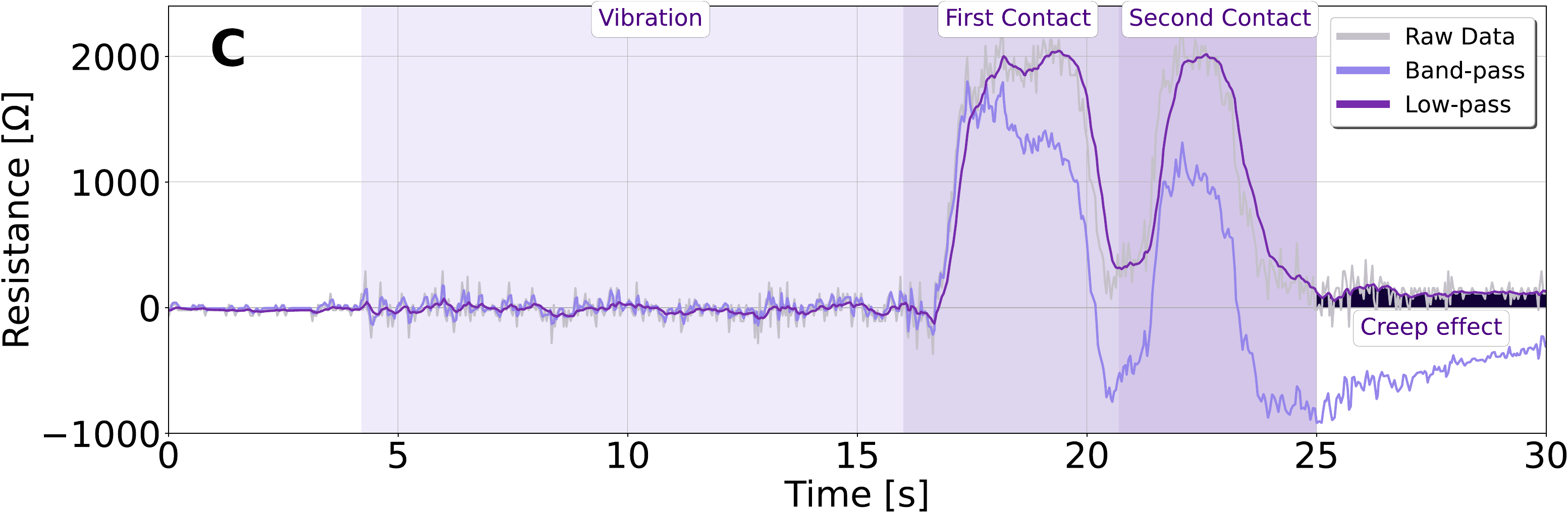}
    % \caption{(a) Whisker characteristic after calibration with Instron Tensile Testing Machine (b) Signal processing of one whisker sensors after interaction with a vertical wall}
    \caption{(A) Contact force response of upper and lower horns. (B) Displacement response of upper and lower horns (C) Signal processing of one horn's signal after interaction with a vertical wall.}
    \label{fig:signalprocessing}
\end{figure}

As shown in Figure~\ref{fig:signalprocessing}A, \ref{fig:signalprocessing}B, the elastic horn sensors demonstrate a nonlinear relationship in both applied force and displacement, and distinct hysteresis is observed between loading and unloading curves. As applied force increases, the resistance of the flex sensor also increases. Upper and lower horns exhibit similar resistance profiles for both applied force and displacement. At the end of each cycle, residual resistance offset reflects horn creep shown in the black highlight colour.

\subsection{Horn System Integration and Signal Processing}\label{sec:systemandsignal}

The horn array is mounted directly on a the frame of our custom aerial platform. Four flex sensors are hardwired through voltage divider circuits to 12-bit ADS1015 analog-to-digital converters, utilising \num{47} \unit{\kilo\ohm} as fixed resistances in the circuits. Resistance variations in the flex sensors generate corresponding voltage changes, measured via the ADS1015 using
\begin{equation}\label{eq:voltagedivider}
 R_\text{flex} = R_\text{fixed}(\frac{V_\text{in} - V_\text{measure}}{V_\text{measure}})
\end{equation}
where $R_\text{flex}$ is the varied resistance of flex sensor, $R_\text{fixed}$ is the fixed resistance, $V_\text{in}$ is the input voltage of the circuit which is \num{3.3} V, and $V_\text{measure}$ is the output voltage from ADC.

%The ADC interfaces with the Raspberry Pi 3b+ (1.4 GHz 64-bit quad-core ARM Cortex-A53 processor) through UART connections, configured for maximum sampling at 860 Hz with I2C communication at 50 Hz. 
Four flex sensors are read through voltage dividers by 12-bit ADS1015 ADCs connected to the Raspberry Pi via I2C. The ADS1015 is configured at up to 860 SPS, sensor values are time-stamped and published in ROS2 at 50 Hz.
%All sensor processing, calibration, and filtering are implemented in Python within a ROS2 framework. 
The Raspberry Pi communicates with the flight controller via UART using MAVROS at 921600 baudrate. 
% MAVROS additionally provides network routing to QGroundControl for real-time system status visualisation and monitoring.

The horn sensors are positioned beneath the motors (Figure~\ref{fig:droneplatform}), causing direct vibration to the flex sensors. Coupled with the inherent noise of flex sensors, nonlinear behaviour, hysteresis and creep effects observed in calibration tests, we proposed a real-time preprocessing method to mitigate these effects. 
%Prior studies in \cite{WhiskerAerial} used a Butterworth bandpass IIR filter (0.03–5.75 Hz) but this underperforms for our system during brief contact events as the passband is too high to adequately preserve slower, large-amplitude features characteristic of our horn sensors. We tune a 0.8 Hz low-pass Butterworth filter to suppress vibration/noise while preserving short contact-induced signal peaks. 
Prior studies in \cite{WhiskerAerial} employed a Butterworth bandpass IIR filter (0.03-5.75 Hz). In our setting, contact events are brief, and the horn response contains both a slow baseline deflection and higher-frequency contact transients. The bandpass attenuates the low-frequency component that carries the large-amplitude deflection. We therefore use a Butterworth low-pass filter with a 0.8 Hz cutoff to reduce high-frequency vibration/noise while retaining the dominant deflection trend produced during contact.

In Figure~\ref{fig:signalprocessing}C, we collected the resistance signal when the drone platform is operating at 50\% throttle bench tests without propellers. Two contact events were performed against a vertical surface, followed by release. While the low-pass IIR filter reduces vibration and noise effectively, the creep remains present in our signal response. Since we focus on the passive behaviour of our horn design, creep effect filtering is beyond the scope of this work.

\section{EXPERIMENTS}\label{sec:experiment}
In this section, the flight experiments of the proposed system are presented and discussed.
We evaluate the system using two experiments: 1) Touch-and-go manoeuvre, 2) Pushing manoeuvre. The first objective is to demonstrate that our novel platform can perform repeated passive bumping behaviour and become stable using the morphology of our horn design. The second objective is to show that the standard attitude controller of the drone attached with our horn design can maintain stability when applying normal contact force to a vertical wall.

\begin{table}[b!]
\caption{Components for our drone platform}
\centering
\renewcommand{\arraystretch}{1.2}
\begin{tabularx}{\columnwidth}{>{\bfseries\RaggedRight}l >{\RaggedRight\arraybackslash}X}
\toprule
Frame & HobbyKing™ Totem Q250  \\
Electronic Speed Controller (ESC) & SpeedyBee F7 V3 BL32 50A 4-in-1 ESC \\
Autopilot & Holybro Kakute H7 Mini V1.5 \\
Firmware & PX4 v1.16.1 \\
Companion Computer & Raspberry Pi 3b+ RAM 1GB (Ubuntu OS server) \\
\bottomrule
\end{tabularx}
\label{tab:droneplatform}
\end{table}

\subsection{Experiment setup}\label{subsec:experimentsetup}
The custom-built quadrotor platform comprises components listed in Table~\ref{tab:droneplatform}. The aircraft has an all-up weight of 700 g. All experiments were conducted with manual remote control (RC) by a human pilot. 
%Control inputs for both experiments were limited to throttle commands for attitude control and pitch inputs for aerial physical interaction manoeuvres. 
Command inputs were limited to throttle for altitude/thrust management and pitch for interaction, roll/yaw inputs were kept near zero.
In the first experiment, the pitch command switches from forward to zero immediately upon wall contact, whereas the second experiment maintained continuous pitch commands throughout contact phases. All data acquisition was managed through the ROS2 framework running on a Raspberry Pi 3B+. The sampling rate of both the onboard IMU of the flight controller and the IMU topic in MAVROS was set to 100 Hz.

% All propeller guards, landing gear, and battery protection cases are 3D-printed using ABS plastic in accordance with the risk assessment standards of Bristol Robotics Laboratory (BRL).

% The aircraft has an all-up weight of 700 g. Both experimental trials were conducted within the indoor flight facility at BRL. Safety mattresses were positioned on the floor as protective measures against potential crashes. As shown in Figure~\ref{fig:experimentsetup}, the drone was positioned facing the vertical wall for the first experiment and oriented toward the panel surface for the second experiment.

%Whisker sensor calibration procedures were completed prior to each experimental run to collect the baseline as described in Section~\ref{sec:whiskercharacteristic}. Following calibration, the aircraft was switched to Stabilised flight mode via RC input. 

\section{RESULTS}\label{sec:results}
\subsection{Touch-and-go manoeuvre}\label{subsec:touchandgo}

% As described in Section~\ref{sec:concept}, 
We conducted experiments mimicking the flight behaviour of houseflies during repeated, brief wall interactions. Following take-off to a specified altitude, the drone was controlled exclusively through pitch angle commands via RC input.% This experimental setup emulates the scenario where a flying insect already extends its legs for landing but fails to establish stable contact or adhere the vertical wall, resulting in a collision.

\begin{figure}[t!]
    \centering
    % \subfloat[The sequence of passive recovery mechanism of the whisker sensors during one collision. (A) The bumper drone is approaching the vertical wall with initial pitch angle of \num{0.2} \unit{\radian}. (B) Two upper whisker sensors first contact with the wall. (C) Due to collision dynamics, two lower whisker sensors contact with the wall. (D) Four Whisker sensors absorb collision energy. (E) The drone bumper starts to bounce-off. (F) The drone is bounced-off the wall and stabilise. This collision occurs within \num{200}\unit{\milli\second}.\label{fig:bumping}]{%
    %     \includegraphics[width=0.85\linewidth]{figures/bumping.png}}
    % \hfill
    
    \includegraphics[width=\linewidth]{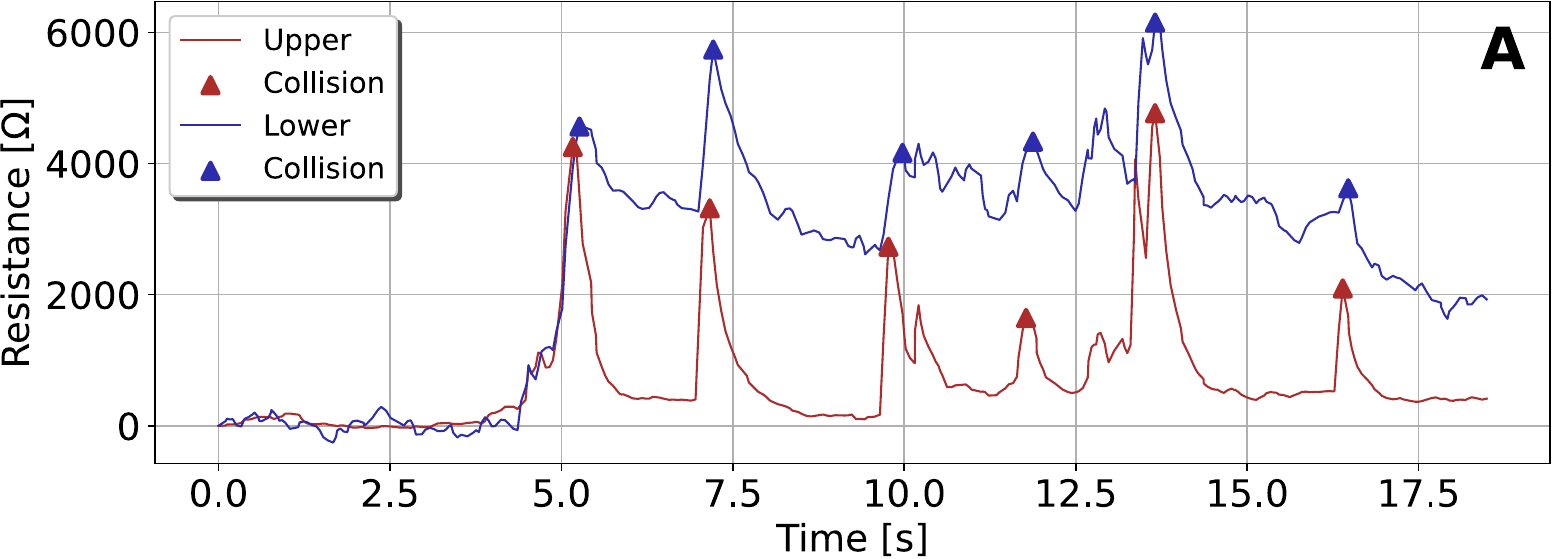}
    \hfill
    \includegraphics[width=\linewidth]{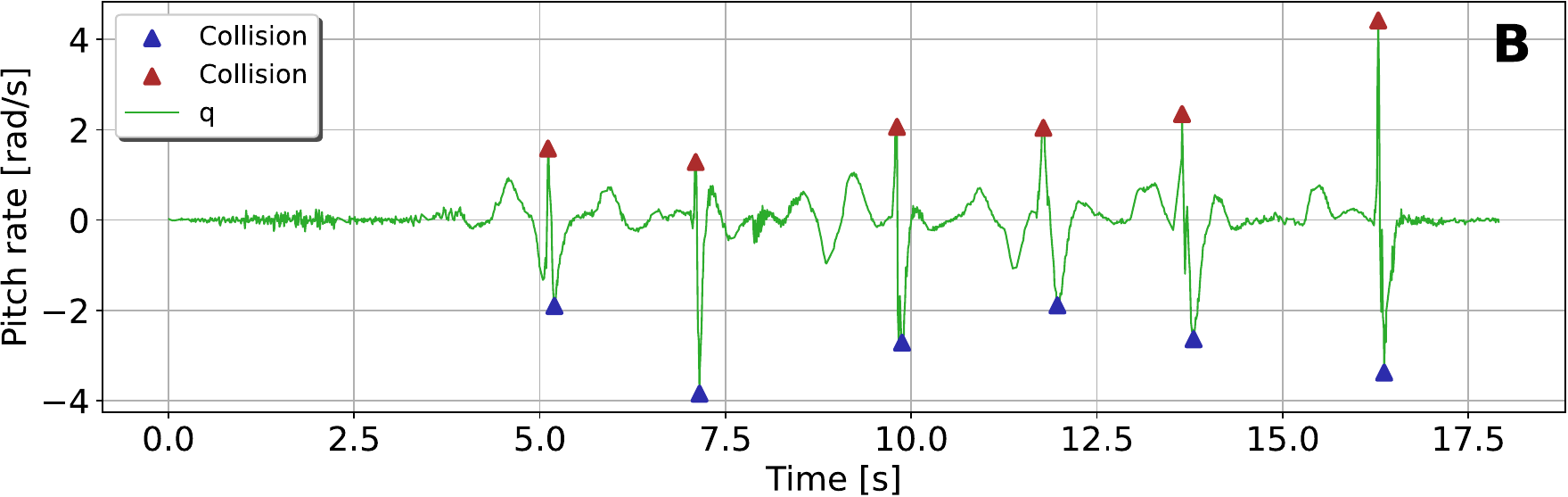}
  \caption{(A) The horn resistance responses of upper horns (red) and lower horns (blue) of one flight experiment. The plot shows that the upper horns make contact first, and the wall generates a positive pitch attitude, causing the drone to rotate, followed by lower horn contact. The total collision count is 6 occurrences. (B) The pitch rate of the drone platform corresponding to the resistance response.}
  \label{fig:bumpingconceptresult} 
\end{figure}

\begin{figure*}[htb]
    \centering
    {\includegraphics[width=0.85\linewidth]{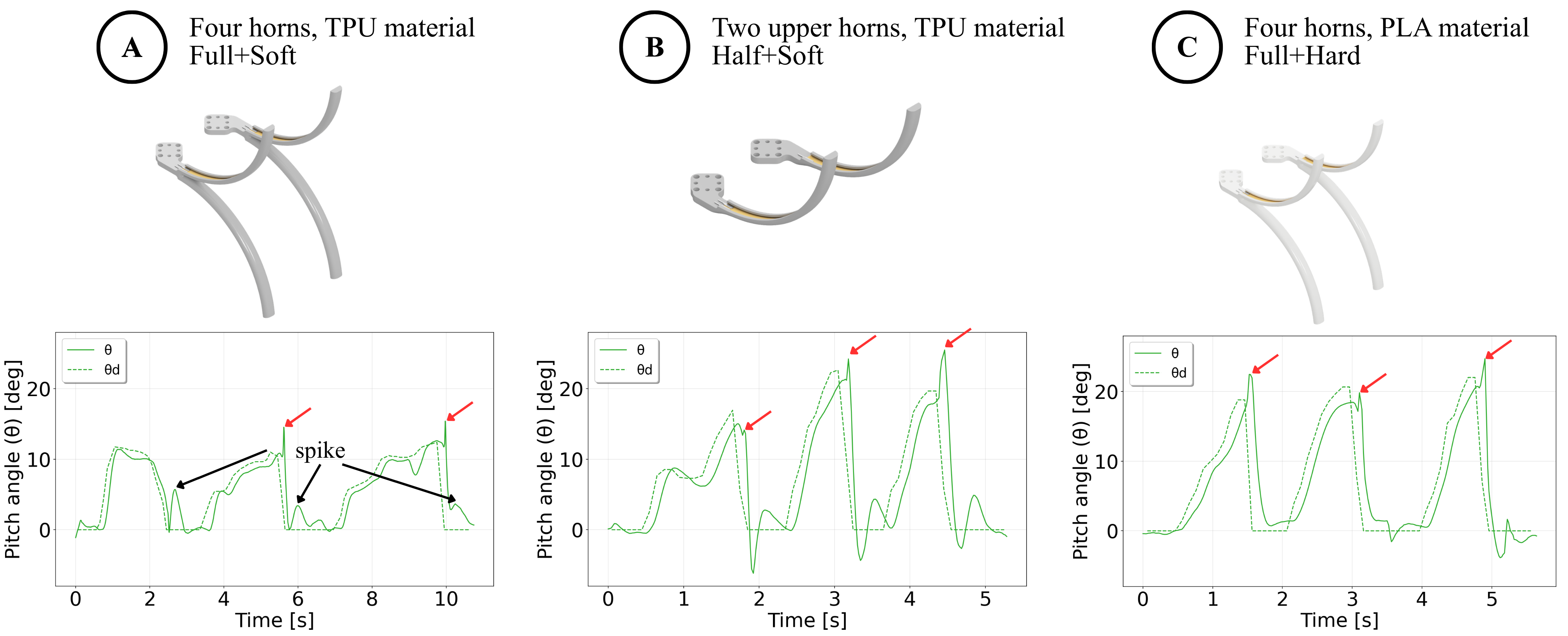}}
    \caption{Pitch attitude response of three configurations. Each configuration experiences three bumping behaviour events. The red arrows indicate the maximum deformation of the upper horns of that collision event.}
  \label{fig:bumpingbehaviours} 
\end{figure*}

Figure~\ref{fig:bumpingconceptresult} illustrates the horn resistance responses and pitch rate of passive bumping behaviour throughout one flight experiment with 6 collisions. Figure~\ref{fig:bumpingconceptresult}A shows the horn resistance reading during the experiment, where the red and blue lines represent the filtered resistance explained in Section~\ref{sec:systemandsignal} of the upper and lower horns, respectively. The triangles indicate maximum resistance of the two horns, corresponding to the contact event. After full deformation of the upper horn, the lower horn passively establishes contact within an average of \num{80}\unit{\milli\second} due to the collision dynamics. The upper horn exhibits sharp resistance spikes, while the lower horn shows more gradual resistance changes that take longer to return to baseline, suggesting a damping role with slower recovery dynamics that help absorb collision impact.

This sequential contact pattern is further revealed by the pitch rate response shown in Figure~\ref{fig:bumpingconceptresult}B. During each collision, initial upper horn contact generates a positive moment, producing an upward pitch rate spike until the upper horns are fully deformed (red triangles), immediately followed by a sharp negative pitch rate as the lower horns establish contact and deform (blue triangles). The negative dip represents the rotational dynamics during collision, with pitch rate gradually returning to neutral as the drone disengages. The positive-then-negative pattern of pitch rate across multiple collision events, combined with the resistance data, demonstrates the passive bumping behaviour achieved by our design.

To assess the influence of horn design on bumping dynamics, three configurations are compared: (1) TPU with a four-horn array (\textit{Full+Soft}); (2) TPU with only two upper horn array (\textit{Half+Soft}); (3) PLA with a four-horn array (\textit{Full+Hard}). A total of 9 additional flight experiments were conducted (three per configuration). Although the soft-horn configurations could sustain a greater number of impacts, each flight was limited to three consecutive bumping events to maintain experimental consistency, as failure of the rigid-horn configuration occurred by the fourth impact. Comparative pitch attitude profiles during three consecutive collision events are presented in Figure~\ref{fig:bumpingbehaviours}.

In both \textit{Full+Soft} and \textit{Full+Hard} configurations, the lower horns store collision energy and provide a restoring moment, resulting in a secondary upward pitch and progressive oscillation damping. In contrast, the \textit{Half+Soft} configuration, which lacks lower horns, induces a negative pitch response due to the wall reaction moment. While \textit{Full+Hard} yields a smooth pitch response, repeated collisions lead to damage, as the rigid horns are unable to absorb impact energy effectively.

\begin{figure}[htb]
    \centering
    \includegraphics[width=0.9\linewidth]{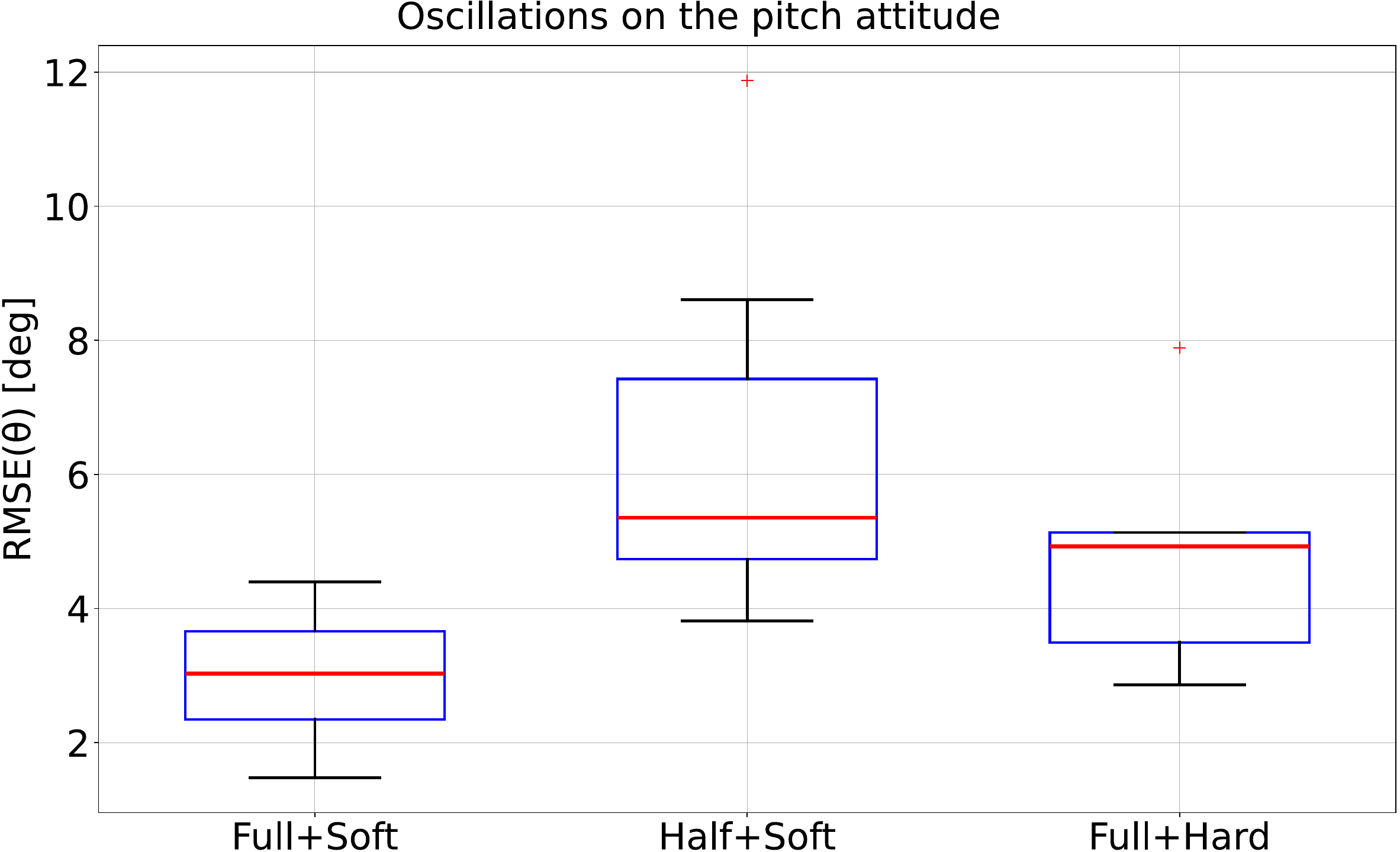}
    \caption{Oscillations in pitch attitude of three configurations: Full+Soft represents two upper horns and two lower horns made of TPU material; Half+Soft represents two upper horns made of TPU material; and Full+Hard represents two upper horns made of PLA material.}
    \label{fig:pitchoscillation}
\end{figure}

\begin{figure}[htb]
    \centering
    \includegraphics[width=\linewidth]{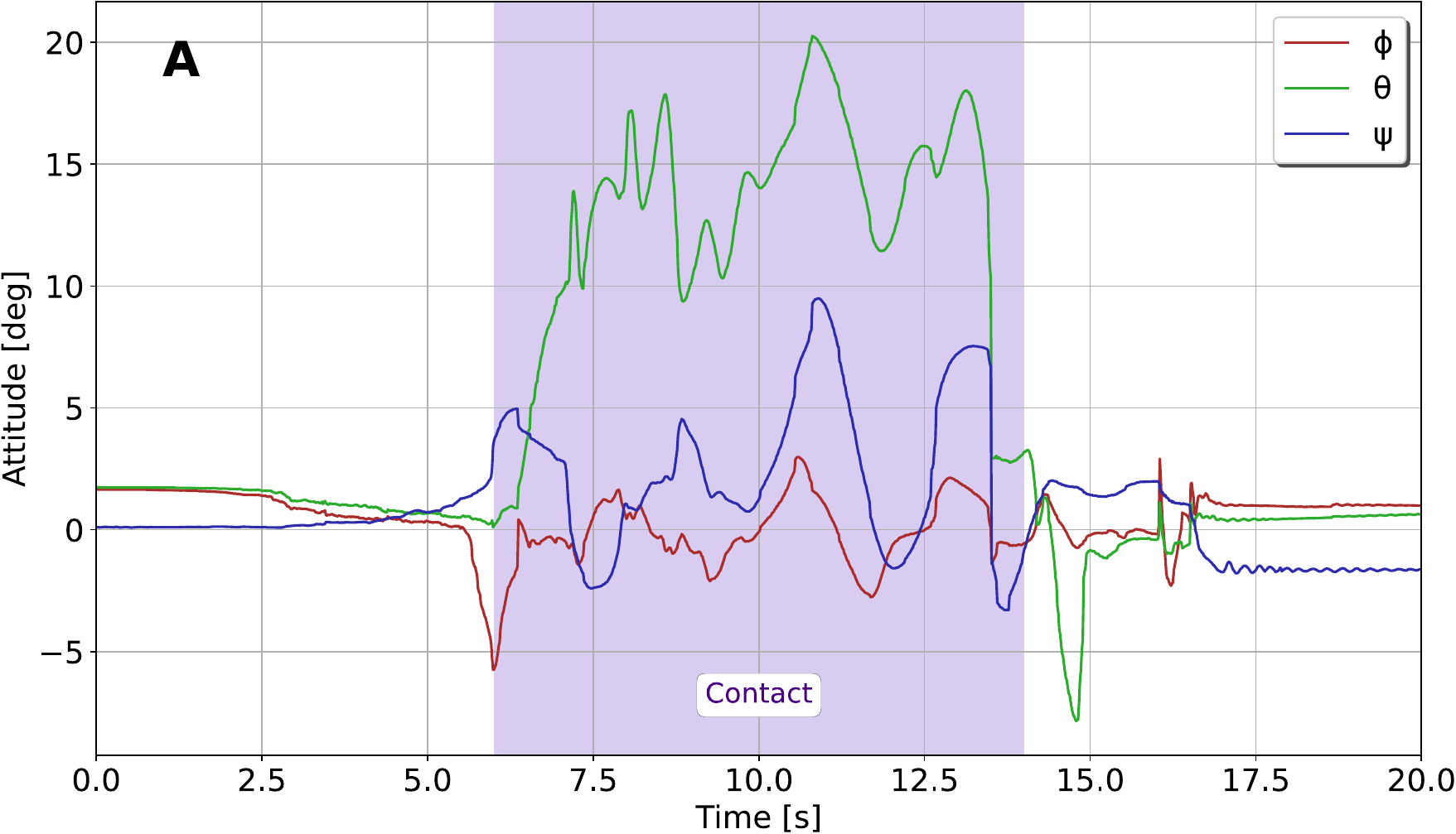}
    \hfill
    \includegraphics[width=\linewidth]{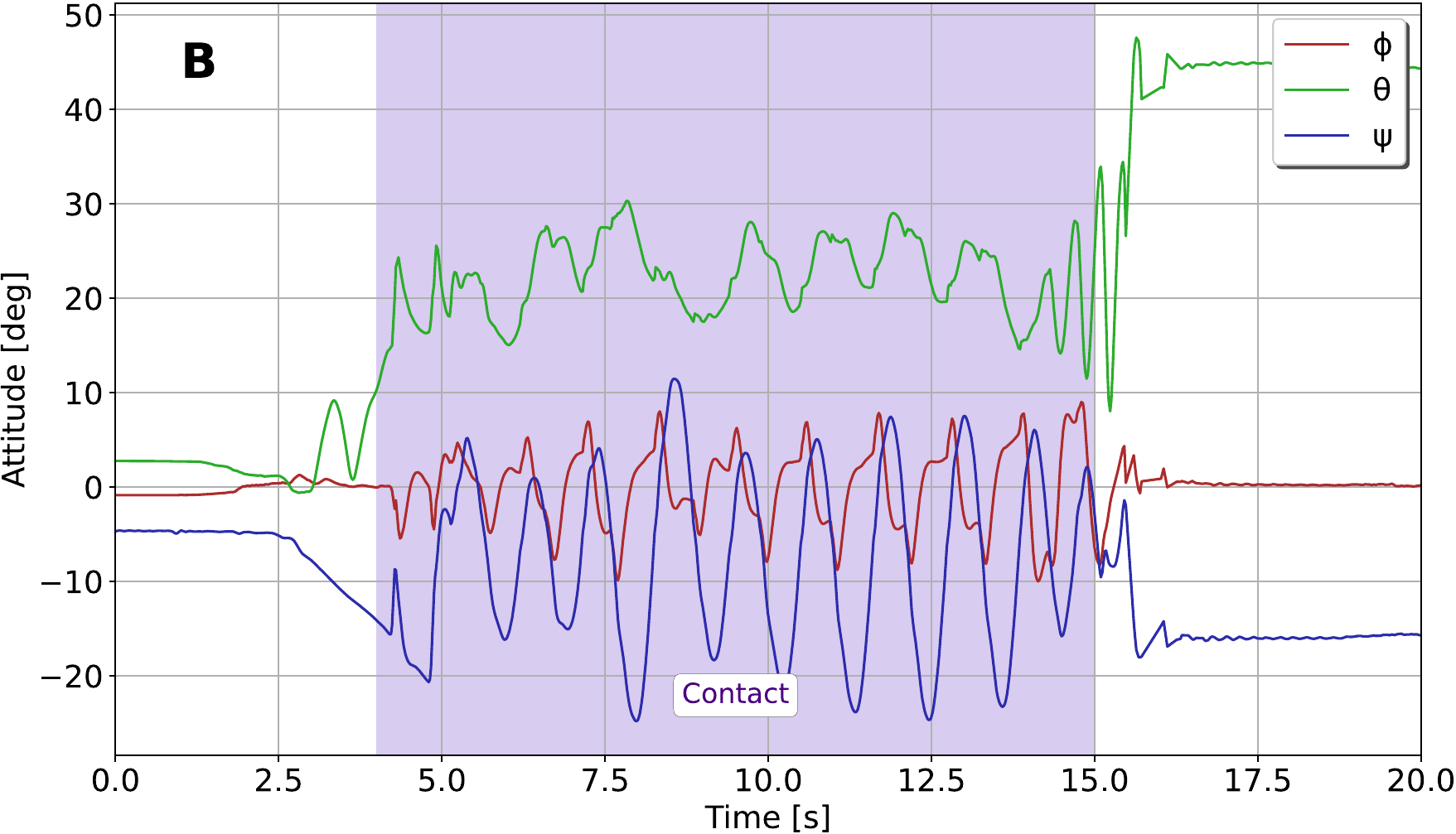}
    \caption{Attitude (roll, pitch, yaw) of the drone platform for the pushing manoeuvre experiment. (A) \textit{Full+Soft} configuration (B) \textit{Full+Hard} configuration.}
  \label{fig:orientationpushing} 
\end{figure}

\begin{figure*}[htb]
    \centering
    \includegraphics[width=\linewidth]{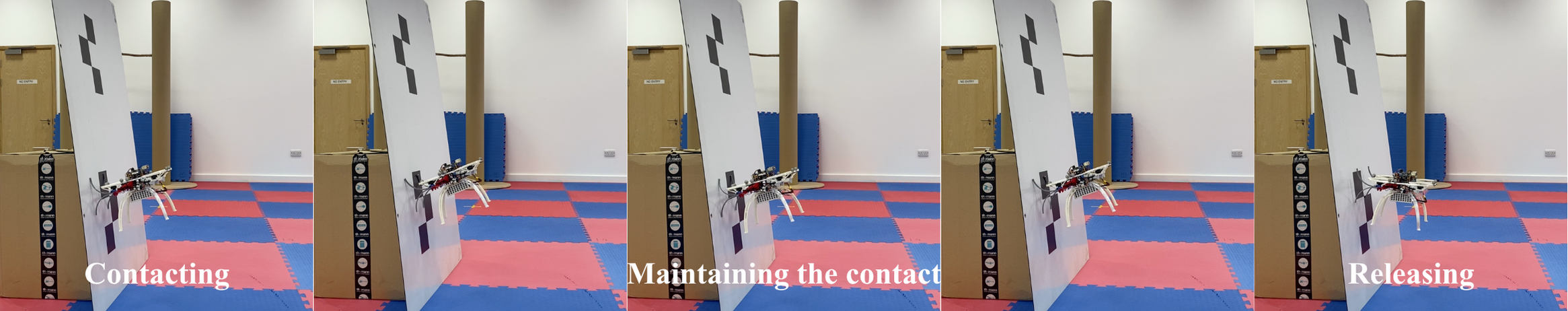}
    \caption{The pushing manoeuvre sequence began with the two upper horns contacting the wall under a pitch angle command from the RC input. The drone was then held at the approaching pitch angle to maintain contact. Subsequently, a negative pitch angle command was applied to bring the lower horns into contact with the wall, serving as the releasing action.}
    \label{fig:processexperiment2}
\end{figure*}

% Moreover, the repeatability of the platform was tested across 17 collisions conducted over 3 flight experiments. Figure~\ref{fig:bumping_profile} plots show the pitch angle, forward velocity $(V_\text{x})$, and forward acceleration $(A_\text{x})$ during collision, averaged across all flight experiments. The shaded regions represent the standard deviation across the 17 collisions. The two vertical dashed lines at $t=\num{0}\unit{\milli\second}$ and $t\approx\num{219}\unit{\milli\second}$ correspond to the start and end of contact event. The temporal responses exhibit characteristics similar to a mass-spring-damper system. The pitch angle (A) demonstrates a smooth exponential decay typical of an overdamped response, while the velocity profile (B) shows the characteristic step response with rapid deceleration followed by gradual recovery. The acceleration trace (C) captures the impulse-like response at impact, with a sharp negative spike. 
% The stability of the drone during collision interactions is assessed by analysing attitude oscillations. The amplitude of oscillations is calculated using the root mean square (RMS) of the roll angle ($\phi$) and pitch angle ($\theta$) as shown in Figure~\ref{fig:pitchoscillation}. The results demonstrate low pitch oscillations showing an RMS value of only \num{0.072}\unit{rad} (\ang{4}) and roll oscillations of just \num{0.013}\unit{rad} (\ang{0.74}).

We analyse the pitch attitude oscillations to assess the stability of the drone with our horn designs during collision events. The amplitudes of oscillations are calculated using the root mean square error (RMSE) of the pitch angle ($\theta$) relative to the setpoint, computed over nine bumping motions per configuration (three flights × three bumps), as shown in Figure~\ref{fig:pitchoscillation}. The means are reported as 3.0 deg, 6.5 deg, and 4.9 deg for the \textit{Full+Soft}, \textit{Half+Soft}, and \textit{Full+Hard} configurations, respectively. The results show that the \textit{Full+Soft} configuration achieves the lowest pitch oscillations, suggesting that elastic material effectively absorbs impact energy and reduces post-collision instability. The plot also reveals that the four-horn array provides better stability than the upper two-horn design, even when using rigid material. This suggests that both material compliance and morphological designs contribute to collision mitigation passively.

% \begin{figure}[htb]
%     \centering
%     \subfloat[\label{fig:orientationpushing0.3}]{%
%         \includegraphics[width=\linewidth]{figures/orientationpushing.png}}
%     \hfill
%     \subfloat[\label{fig:orientationpushing0.8}]{%
%         \includegraphics[width=\linewidth]{figures/orientationpushing2.png}}
%     % \hfill
%     % \subfloat[Orientation of UAV without I gain from rate controller\label{fig:orientationpushingI}]{%
%     %     \includegraphics[width=\linewidth]{figures/orientationpushing3.png}}
%     \caption{Orientation of UAV in the static pushing task experiment. The yaw angle oscillates in a small region and even oscillates more if the pitch angle increases, corresponding to an increase in contact force.}
%   \label{fig:orientationpushing} 
% \end{figure}

\subsection{Pushing manoeuvre}\label{subsec:pushing}
The pushing manoeuvre experiments were conducted in three phases as illustrated in Figure~\ref{fig:processexperiment2}: target approach and contact, sustained contact, and contact release. The drone was commanded using only forward and backward pitch inputs via RC control and depended on a PID-based attitude controller.

Figure~\ref{fig:orientationpushing}A presents the attitude response of the drone platform equipped with TPU horns during the pushing manoeuvre experiment. The pitch input was held constant at approximately 15 deg, and the drone maintained contact with the pitch attitude oscillating between 10 deg and 20 deg. During sustained contact, roll and yaw oscillations remained contained within –3 deg to 3 deg and –8.5 deg to 11 deg, respectively, indicating that while the horn design effectively manages pitch-direction forces, some lateral disturbances from contact are still present.

A comparison with PLA horns (Figure~\ref{fig:orientationpushing}B) shows that, under a constant pitch input around 20 deg, the pitch attitude oscillated in a similar range (15 deg - 30 deg) to the TPU case, but roll and yaw disturbances were significantly greater. The roll attitude oscillated from –5 deg to 5 deg, and the yaw attitude from –20 deg to 10 deg. These oscillations grew over the contact duration, resulting in instability, $t\approx$15 s. This experiment highlights that our elastic horn design provides passive disturbance rejection with the regular attitude controller, while the rigid PLA horns transmit more force to the airframe, leading to greater instability and loss of control.

\section{CONCLUSIONS}

In this work, we proposed a novel morphology design to handle uncertain contact forces by utilising touch-and-go manoeuvre to reduce the reliance on active control during aerial pushing and/or sliding tasks. An elastic horn design was developed to store impact energy and behave as a mass–spring–damper system, resulting in passive bumping behaviour with unknown surfaces. The elasticity and morphology of the horns facilitate stable, consecutive bumping as a form of embodied control. Experimental results demonstrated that the elastic horn improves vehicle stability, reducing pitch oscillations by 38\% compared to the rigid horn configuration, while a combined upper–lower horn configuration reduces pitch oscillations by 54\%. Furthermore, the platform demonstrated stable pushing manoeuvres, maintaining contact with vertical walls with minimal control effort.

To better understand the underlying interaction dynamics, future work will also focus on developing a dynamic model of the drone–horn–environment system. In particular, we aim to characterise how impact energy is dissipated through the elastic morphology and how effective mass–spring–damper behaviour emerges during contact. 

In addition, tactile sensing will be integrated into closed-loop control for autonomous touch-and-go manoeuvres, allowing the drone to actively adjust its attitude for the subsequent bumping actions across a range of unknown surfaces. The stabilisation of the drone is expected to be improved with tactile feedback control for pushing manoeuvres. 

Since our experiments demonstrated the benefits of integrating morphology and soft materials for aerial physical interaction tasks, we will use this closed-loop motion controller to benchmark against state-of-the-art force-control strategies, such as impedance or admittance control. This comparison will clarify whether the passive mechanical properties of the system can reduce the need for force control and accurate force sensing during contact-rich aerial interactions.

\addtolength{\textheight}{-6cm}   % This command serves to balance the column lengths

%%%%%%%%%%%%%%%%%%%%%%%%%%%%%%%%%%%%%%%%%%%%%%%%%%%%%%%%%%%%%%%%%%%%%%%%%%%%%%%%

%%%%%%%%%%%%%%%%%%%%%%%%%%%%%%%%%%%%%%%%%%%%%%%%%%%%%%%%%%%%%%%%%%%%%%%%%%%%%%%%



\section*{ACKNOWLEDGMENT}

We would like to thank Bristol Robotics Laboratory for providing the flight testing facility and also acknowledge the experimental support provided by technician Patrick Brinson. Furthermore, we acknowledge Ziniu Wu, Ziang Zhao, and Haichuan Li for their contributions in hardware design and flight test support.

%%%%%%%%%%%%%%%%%%%%%%%%%%%%%%%%%%%%%%%%%%%%%%%%%%%%%%%%%%%%%%%%%%%%%%%%%%%%%%%%

% References are important to the reader; therefore, each citation must be complete and correct. If at all possible, references should be commonly available publications.

\bibliographystyle{IEEEtran}
\bibliography{References}

\end{document}